%% file: main.tex
\documentclass{article}

\usepackage[preprint]{corl_2026} 
\makeatletter
\@ifpackagewith{corl_2026}{preprint}{
  \newcommand{\projecturl}{\href{https://amrl.cs.utexas.edu/foresight}{https://amrl.cs.utexas.edu/foresight}}
}{
  \@ifpackagewith{corl_2026}{final}{
    \newcommand{\projecturl}{\href{https://amrl.cs.utexas.edu/foresight}{https://amrl.cs.utexas.edu/foresight}}
  }{
    \newcommand{\projecturl}{\href{https://foresightplanner.github.io/foresight}{https://foresightplanner.github.io/foresight}}
  }
}
\makeatother

\title{Foresight: Iterative Reasoning About Clues that Matter for Navigation}

\author{
Arthur Zhang$^{1,\dagger}$,
Carl Qi$^{1,\dagger}$,
Donne Su$^{1}$,
Xiangyun Meng$^{2}$,
Amy Zhang$^{1}$,
Joydeep Biswas$^{1}$\\
$^{1}$UT Austin, $^{2}$FieldAI
}
\input{definitions}
\begin{document}
\maketitle

\begingroup
\renewcommand{\thefootnote}{\ensuremath{\dagger}}
\footnotetext{Corresponding authors: \texttt{\{arthurz,carlqi\}@cs.utexas.edu}.}
\endgroup

\vspace{-1.5em}

\begin{abstract}
Open-world mapless navigation from sparse language instructions requires resolving underspecified goals and inferring which environmental cues are relevant for reaching the goal. For instance, reaching an out-of-view destination may require interpreting ramps, signs, or detours that reveal where to go or which route to take. Prior works are limited by their reliance on known navigation factors and closed-set factor categories, or identify cues before motion planning and miss plan-dependent cues. We argue that pretrained Vision-Language Models (VLMs) can discover novel instruction-relevant cues, but require adaptation to focus on which cues matter and how they should influence motion planning. We realize these ideas in \ourmodel{}, a test-time framework in which a finetuned VLM alternates between proposing image-space motion plans and critiquing them using the language goal and visual context. Subsequent plans are conditioned on prior critiques, enabling iterative motion refinement before execution. To align plan critiques and refinements with open-set behavior preferences, we learn a reward model from human feedback and use it to post-train the VLM with reinforcement learning in the plan-critique loop. In offline evaluations and 6 real-world environments, \ourmodel{} improves average task success by 37\% and reduces interventions per mission by 52\% relative to state-of-the-art test-time reasoning and foundation-model baselines, while running in real-time on a Jetson AGX Orin. We will release code, data, and training details to support future work on test-time reasoning for robot motion refinement. Additional videos at: \projecturl

\end{abstract}



\input{introduction}

\input{relatedwork}

\input{preliminaries}
\input{approach}
\input{experiments}

\input{conclusion}

\input{limitations}







\clearpage
\acknowledgments{
This work has taken place in the Autonomous Mobile Robotics Laboratory (AMRL) and Machine Decision-making through Interaction Laboratory (MIDI) at UT Austin. AMRL research is supported in part by NSF (CAREER-2046955, IIS-2416461,  PARTNER-2402650), ARO (W911NF-24-2-0025), and by FieldAI. MIDI research is supported in part by NSF (CAREER-2340651, PARTNER-2402650), DARPA (HR00112490431), and ARO (W911NF-24-1-0193). We thank Arjun Guha and Amirreza Shaban for their support, advice, comments, and discussions during the project. Any opinions, findings, and conclusions expressed in this material are those of the authors and do not necessarily reflect the views of the sponsors.
}


\bibliography{refs}  

\input{appendix}

\end{document}

%% file: definitions.tex
\usepackage{amsmath, amssymb, amsthm}
\usepackage{mathtools}
\usepackage{comment}
\usepackage{caption}
\usepackage{makecell}
\usepackage{diagbox}
\usepackage{wrapfig}
\usepackage{graphicx}
\usepackage{multirow}
\usepackage{enumitem}
\usepackage{booktabs}

\usepackage[most]{tcolorbox}
\usepackage{fvextra}

\newcommand{\ourmodel}{\textsc{Foresight}}

\newcommand{\amy}[1]{[{\textbf{\textcolor{blue}{Amy: #1}}}]}
\newcommand{\akz}[1]{[{\textbf{\textcolor{magenta}{Arthur: #1}}}]}
\newcommand{\CQ}[1]{[{\textbf{\textcolor{cyan}{Carl: #1}}}]}

\newcommand{\seclabel}[1]{\label{sec:#1}}
\newcommand{\secref}[1]{Sec.~\ref{sec:#1}}
\newcommand{\sseclabel}[1]{\label{ssec:#1}}
\newcommand{\ssecref}[1]{Sec.~\ref{ssec:#1}}

\newcommand{\figlabel}[1]{\label{fig:#1}}
\newcommand{\figref}[1]{Fig.~\ref{fig:#1}}
\newcommand{\tablabel}[1]{\label{tab:#1}}
\newcommand{\tabref}[1]{Table~\ref{tab:#1}}
\newcommand{\eqnlabel}[1]{\label{eq:#1}}
\newcommand{\eqnref}[1]{Eq.~\ref{eq:#1}}

\newcommand{\KL}{\mathrm{KL}}
\newcommand{\IG}{\mathrm{IG}}

\newcommand{\E}{\mathbb{E}}
\newcommand{\refp}{\pi^{\mathrm{refine}}_\theta}
\newcommand{\planp}{\pi^{\mathrm{plan}}_\theta}

\newcommand{\dr}{\overline{\Delta r}}

\makeatletter
\renewcommand{\section}{%
  \@startsection{section}{1}{\z@}%
                {-0.6ex \@plus -0.2ex \@minus -0.1ex}%
                {0.15ex \@plus 0.05ex}%
                {\large\bf\raggedright}%
}
\renewcommand{\subsection}{%
  \@startsection{subsection}{2}{\z@}%
                {-0.45ex \@plus -0.15ex \@minus -0.1ex}%
                {0.10ex \@plus 0.05ex}%
                {\normalsize\bf\raggedright}%
}
\renewcommand{\subsubsection}{%
  \@startsection{subsubsection}{3}{\z@}%
                {-0.35ex \@plus -0.1ex \@minus -0.05ex}%
                {0.05ex \@plus 0.03ex}%
                {\normalsize\bf\raggedright}%
}
\makeatother

\definecolor{fsPromptBg}{RGB}{248,248,248}
\definecolor{fsPromptFrame}{RGB}{210,210,210}
\definecolor{fsPlaceholderBlue}{RGB}{35,90,160}

\newtcolorbox{promptbox}[1][]{
  colback=fsPromptBg,
  colframe=fsPromptFrame,
  boxrule=0.4pt,
  arc=1.5pt,
  left=5pt,
  right=5pt,
  top=5pt,
  bottom=5pt,
  breakable,
  enhanced,
  #1
}

\DefineVerbatimEnvironment{PromptVerbatim}{Verbatim}{
  fontsize=\scriptsize,
  breaklines=true,
  breaksymbolleft={},
  breaksymbolright={},
  breakindent=0pt,
  breakautoindent=false,
  tabsize=2
}

%% file: introduction.tex
\section{Introduction}
\seclabel{introduction}

Deploying general-purpose robots in everyday environments requires navigation systems that can follow sparse instructions without high-definition maps, predefined routes, or exhaustive task specifications. This is challenging because successful navigation often depends on unknown open-set visual cues whose relevance is determined by the goal and local context. For example, reaching a building entrance may require interpreting signs, ramps, or badge readers that distinguish the correct route from visually similar alternatives. Such cues may be absent from training data and difficult to encode with predefined semantic classes or rules.

Learning from Demonstration (LfD) offers a potential path toward open-world navigation, but demonstrations provide weak supervision for identifying which visual cues matter and how they should influence routing decisions. Feed-forward policies~\cite{zhang2025creste, zitkovich2023rt, hirose2025omnivla, hiroselelan} learn direct observation-to-action mappings that implicitly treat familiar and novel factors equally, limiting understanding of novel cues that cannot be directly matched to prior demonstrations. Pre-emptive reasoning methods~\cite{zawalski2025robotic, zhao2025cot, wang2025alpamayo, intelligence2025pi_} infer task-relevant factors before planning, but cue relevance is often plan-dependent: a detour matters only if the current plan follows a blocked path. Iterative methods~\cite{kim2024pre, han2024interpret, huang2023inner} address this by evaluating and refining plans, but existing approaches rely on human critiques~\cite{huang2023inner} or symbolic plans~\cite{kim2024pre, han2024interpret}, limiting their applicability to continuous open-world navigation.

\input{figures/mainfigure}

In this paper, we present \ourmodel{}, an iterative refinement framework and scalable training recipe for continuous robot motion planning. \ourmodel{} adapts a pretrained Vision-Language Model (VLM) into a navigation policy that acts as both planner and critic. It proposes an image-space motion plan, critiques the plan with respect to the goal and scene, and uses the critique to refine the next plan until acceptance or a fixed refinement budget is reached. This loop leverages the ability of pretrained VLMs to identify open-world visual cues and reason about their implications for planning, but effective refinement requires adapting the model to focus on navigation-relevant cues and translate critiques into concrete plan updates. We first use supervised finetuning to teach the policy the structure of plan-critique refinement. However, imitation alone provides limited supervision for multi-step refinement because the space of possible plans, critiques, and updates branches rapidly. Reinforcement learning can improve this iterative planning process with outcome-level supervision, but requires rewards that are difficult to hand-design for open-world navigation. We therefore learn a preference reward model from human feedback and use it to post-train the VLM policy, enabling \ourmodel{} to improve its plans, critiques, and refinements without dense ground-truth annotations.

We demonstrate the effectiveness of \ourmodel{} through offline and real-world robot experiments on the task of mapless navigation with sparse language goals. 
In six real-world environments, \ourmodel{} \textbf{improves average task success by 37\%} and \textbf{reduces interventions per mission by 52\%} relative to state-of-the-art test-time reasoning and robotics foundation-model baselines. Notably, these gains come from short, free-form reasoning traces generated by a VLM policy that \textbf{runs in real-time on a Jetson AGX Orin}, outperforming baselines that use larger models, more elaborate reasoning traces, or significantly more training data. Our main contributions are threefold: 1) we formulate iterative plan-critique refinement as a test-time reasoning framework for continuous robot motion planning; 2) we introduce a scalable training recipe that combines supervised finetuning with reinforcement learning from human preferences to adapt VLMs for iterative refinement; and 3) we demonstrate consistent improvements in task success and intervention rate across offline evaluations and closed-loop experiments in six real-world environments.

%% file: figures/mainfigure.tex
\begin{figure}[htbp]
    \centering
    \includegraphics[width=\linewidth]{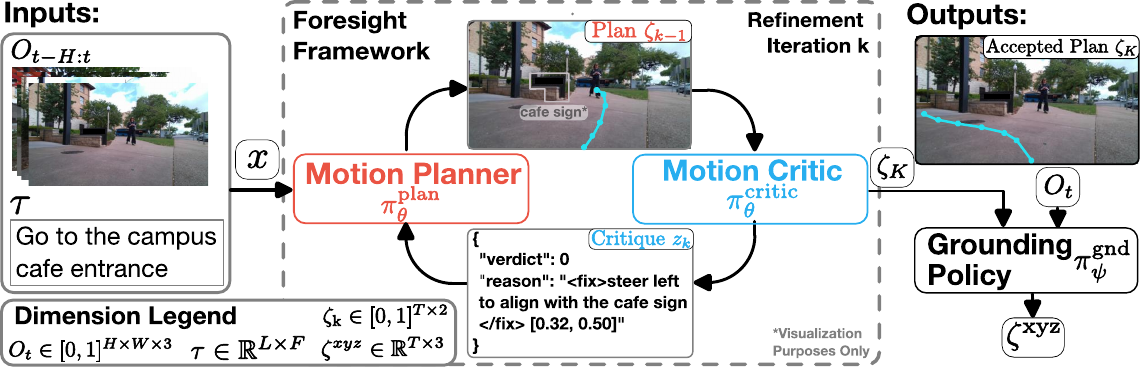}
    \caption{Overview of \ourmodel{} framework. Given image observations $o_{t-H:t}$ and language task $\tau$, \ourmodel{} alternates between generating image space plans $\zeta_{k-1}$ and textual critiques $z_k$, conditioning on prior plan-critique pairs to refine the motion plan. A lightweight grounding policy $\pi^\text{gnd}_\psi$ conditions on the current observation $o_t$ to ground the final plan $\zeta_K$ to a cartesian trajectory.}
    \figlabel{mainfigure}
\end{figure}

%% file: relatedwork.tex
\section{Related Work}
\seclabel{related work}

Language-conditioned mapless navigation requires translating visual observations and sparse language goals into actions while inferring open-world cues whose relevance is unknown a priori. Feedforward policies adapt pretrained representations or VLMs to map observation-language pairs directly to actions~\cite{zhang2026ventura, hirose2025omnivla, hiroselelan, zhang2025creste}, but can struggle when successful navigation depends on cues or cue-action relationships that are under represented in demonstrations. This motivates test-time reasoning methods that allocate additional computation to identify relevant factors, evaluate candidate behaviors, and revise plans before execution.

\subsection{Test-time Reasoning for Robotics}

Pre-emptive approaches are a test-time reasoning method that use foundation models like VLMs to identify relevant scene factors before acting. This approach improves open-set cue discovery, but relies on structured prompts or predefined factor categories~\cite{zawalski2025robotic, wang2025alpamayo}, limiting performance when the relevant cues are unknown a priori. Moreover, deciding which cues matter requires considering how they affect candidate plans. Iterative methods address this plan-dependence by using feedback to revise plans at test time~\cite{huang2023inner} and foundation models to explain failures and refine behavior~\cite{qiself, kim2024pre, han2024interpret}, but typically rely on human/environment feedback or operate over symbolic plans like domain-specific languages and code. Our work follows the iterative refinement strategy, but differs by automatically generating free-form textual critiques with a finetuned VLM and using them to refine continuous motion plans. Furthermore, \ourmodel{} does not require pre-defining domain-specific factor classes, allowing our approach to express open-set cue-plan relationships during refinement.

\subsection{RL Post-training for Robotics}

Beyond test-time reasoning techniques, our work builds on RL post-training methods that adapt pretrained foundation models for robotics tasks. Prior works finetune VLAs from sparse task rewards~\cite{hu2025flare, li2025simplevla}, shaped rewards such as geometric progress~\cite{zeng2025poliformer}, VLM-based progress estimates~\cite{intelligence2025pi}, world-model rollouts~\cite{he2025seeing, yong2026generalizable, li2025vla}, or preference signals from rollouts and human interventions~\cite{zhang2024grape, xia2026human}. Our approach is closest to preference-based post-training as we use preference rewards to adapt a VLM inside an iterative plan-critique-refinement loop, jointly optimizing critiques and motion refinements rather than only aligning with the expert plan. This adaptation is important because pretrained VLMs may recognize open-world cues, but still need to learn which cues are relevant for motion planning and how to translate them into plan improvements. We further combine preference feedback with expert-plan alignment, reflecting that refinements must be both semantically aligned with open-world cues and geometrically consistent with expert robot motion.

%% file: preliminaries.tex
\section{Problem Definition and Background}
\seclabel{preliminaries}

We study open-world navigation with sparse language guidance, where the robot receives a context $x=(o_{t-H:t}, \tau)$ consisting of an image observation history $o_{t-H:t}$ and a natural language instruction $\tau$. Starting at timestep $t$, the robot must execute a Cartesian motion plan $\zeta^{\mathrm{xyz}} = a_{t:t+N}$, where $N$ is the planning horizon and each waypoint $a_t \in \mathbb{R}^3$ specifies a position in the world\footnote{We consider 3D positions for a ground robot to accommodate stairs, ramps, and other non-planar terrains}. Following common practice in vision-based control~\cite{zhang2026ventura, lee2025molmoact}, we represent the planned motion using normalized image-space waypoints $\zeta \in [0,1]^{N \times 2}$, which provide a compact visual specification of the desired path and are grounded into Cartesian waypoints for execution. The objective is to make progress toward the intended goal while respecting the semantic and geometric constraints of the scene. This setting is challenging because sparse instructions may under-specify the goal and route, requiring the robot to infer relevant open-set visual factors and how they influence the planned path.

\textbf{Chain-of-Thought Reasoning} To promote expressive reasoning about novel factors during planning, Chain-of-Thought (CoT) \cite{wei2022chain} introduces intermediate reasoning traces $z$ between the input context $x$ and final trajectory $\zeta$. Instead of modeling the trajectory distribution as $p(\zeta \mid x)$, CoT explicitly generates a reasoning trace $z$ before producing the trajectory:
\begin{equation}
p(\zeta, z \mid x)
= p(\zeta \mid z, x) p(z \mid x)
\end{equation}
For navigation, $z$ encodes route-relevant information such as  subgoals, relevant landmarks, spatial constraints, or high-level task representations, while $\zeta$ specifies the resulting motion plan.

\paragraph{Group Relative Policy Optimization.}
Group Relative Policy Optimization (GRPO)~\cite{shao2024deepseekmath} is a policy-gradient method for optimizing a model from reward feedback. 
Let $\pi_\theta(y \mid x)$ be a policy that generates an output $y$ conditioned on input $x$, and let $\pi_{\theta_{\mathrm{old}}}$ denote the policy used to sample training outputs. For each input $x$, GRPO samples a group of $G$ outputs
$\{y^i\}_{i=1}^{G}$ from $\pi_{\theta_{\mathrm{old}}}(\cdot \mid x)$. Each output receives a scalar reward $r^i$. The rewards are normalized within the group to produce relative advantages
$
    A^i =
\frac{
r^i-\mathrm{mean}_{j}(r^j)
}{
\mathrm{std}_{j}(r^j)+\epsilon
}
$
, where $\epsilon$ is a small constant for numerical stability. 
The normalized advantage $A^i$ measures whether output $y^i$ is better or worse than other samples for the same input, avoiding the need to learn a separate value function.
The policy is then updated via:
\begin{equation}
\mathcal{J}_{\mathrm{GRPO}}(\theta)
=
\E_{x,\{y^i\}_{i=1}^{G}}
\left[
\frac{1}{G}\sum_{i=1}^{G}
\frac{
\pi_\theta(y^i \mid x)
}{
\pi_{\theta_{\mathrm{old}}}(y^i \mid x)
}
A^i
\right]
-
\beta\,
\KL\!\left(
\pi_\theta(\cdot \mid x)
\,\|\, 
\pi_{\mathrm{ref}}(\cdot \mid x)
\right)
\end{equation}
where $\beta$ controls regularization toward a reference policy $\pi_{\mathrm{ref}}$.

%% file: approach.tex
\section{Approach}

\input{figures/trainingrecipe}

We introduce \ourmodel{}, a test-time motion refinement framework for open-world mapless navigation from sparse language instructions. As shown in~\figref{mainfigure}, \ourmodel{} consists of two high-level components: an image-space refinement policy and a lightweight grounding policy. First, a single VLM with role-specific prompts samples an iterative plan-critique trace $p_\theta(\zeta_{0:K}, z_{0:K-1} \mid x)$ and produces a refined image-space motion plan $\zeta_K$. Second, a lightweight transformer-based grounding policy $\pi^\text{gnd}_\psi$ maps $\zeta_K$ and current observation $o_t$ to a metric trajectory $\zeta^{\mathrm{xyz}}$. Together, this defines the following factorization:
\begin{equation}
\eqnlabel{full-policy-factorization}
p_{\theta,\psi}(\zeta^{\mathrm{xyz}}, \zeta_{0:K}, z_{0:K-1} \mid x, o_t)
=
\underbrace{
\pi^\text{gnd}_\psi(\zeta^{\mathrm{xyz}} \mid o_t, \zeta_K)
}_{\text{grounding policy}}
\,
\underbrace{
p_\theta(\zeta_{0:K}, z_{0:K-1} \mid x)
}_{\text{\ourmodel{} trace}} .
\end{equation}
\eqnref{full-policy-factorization} relies on \ourmodel{} to accurately steer the grounding policy to generate safe, task-aligned executable motion trajectories. We provide implementation details for $\pi^\text{gnd}_\psi$ in the Appendix~\ssecref{appendix:trainingdetailsgrounding}. In the remainder of this section, we formalize iterative motion refinement as a chain-of-thought factorization, describe the SFT pre-training procedure, and present the reward design and RL post-training recipe for the plan-critique loop.

\subsection{Iterative Motion Refinement as Chain-of-Thought Planning}

Let $x=(o_{t-H:t}, \tau)$ denote the navigation context, consisting of an observation history and sparse language instruction. We cast motion refinement as an iterative CoT process indexed by refinement step $k=0,\ldots,K$, distinct from the robot execution timestep $t$. A single VLM policy $\pi_\theta$ uses role-specific prompts to define a planner $\pi^\text{plan}_\theta$ and critic $\pi^\text{critic}_\theta$, where the planner generates an initial motion plan $\zeta_0$ from $x$, the critic generates a textual critique trace $z_k$ that evaluates the current plan $\zeta_k$, and the planner generates the next plan $\zeta_{k+1}$ conditioned on both $\zeta_k$ and $z_k$:
\begin{equation}
\eqnlabel{iterative-cot}
p_\theta(\zeta_{0:K}, z_{0:K-1} \mid x)
=
\underbrace{\pi^{\mathrm{plan}}_\theta(\zeta_0 \mid x)}_{\text{initial plan}}
\prod_{k=0}^{K-1}
\underbrace{\pi^{\mathrm{plan}}_\theta(\zeta_{k+1} \mid \zeta_k, z_k, x)}_{\text{critique-conditioned plan}}
\,
\underbrace{\pi^{\mathrm{critic}}_\theta(z_k \mid \zeta_k, x)}_{\text{critique}} .
\end{equation}
Here, each critique $z_k$ provides textual feedback about visual cues, task constraints, or plan corrections relevant to improving $\zeta_k$. This factorization couples critique generation with planning, encouraging the critic to produce refinement-relevant feedback and the planner to use critique-identified factors to adapt motion plans to cues not specified a priori or seen in demonstrations.

\subsection{Supervised Pre-training} 
\sseclabel{approach:sft}

We begin with supervised finetuning to warm-start a policy for the proposed plan-critic roles in~\eqnref{iterative-cot}. As shown in~\figref{trainingrecipe}, we first train a base planner $\pi^\text{motion}_\theta$ to imitate expert plans $\hat{\zeta}$ and then sample noisy candidate plans $\zeta\sim\pi^\text{motion}_\theta(x)$. We use an Gemini-3.1-Flash~\cite{team2025gemini} to generate oracle critiques $\hat{z}$ that describe the relevant factors and how to improve $\zeta$ to align with the task instruction. This yields single-step refinement tuples $\mathcal{D}_{\mathrm{SFT}} = \{(x,\zeta,\hat{z},\hat{\zeta})\}$, where $\hat{\zeta}$ denotes the expert motion plan. We supervise the critic $\pi^\text{critic}$ to predict the oracle critique and the planner in two contexts: predicting the expert plan directly from $x$, and predicting the expert plan conditioned on a noisy plan and critique. The SFT objective is
\begin{equation}
\label{eq:sft-loss}
\mathcal{L}_{\mathrm{SFT}}(\theta)
=
-\mathbb{E}_{(x,\zeta,\hat{z},\hat{\zeta}) \sim \mathcal{D}_{\mathrm{SFT}}}
\left[
\log \pi^{\mathrm{plan}}_\theta(\hat{\zeta} \mid x)
+
\log \pi^{\mathrm{critic}}_\theta(\hat{z} \mid \zeta, x)
+
\log \pi^{\mathrm{plan}}_\theta(\hat{\zeta} \mid \zeta, \hat{z}, x)
\right].
\end{equation}

This warm-start provides high-quality offline supervision, but pairing oracle critiques of varying relevance with the same expert target plan prevents SFT from learning which critiques best guide refinement. We address this limitation using reinforcement learning to jointly optimize critique generation and critique-conditioned motion planning.


\subsection{Preference-Based RL Post-Training}
\sseclabel{approach:rl}

We post-train the SFT policy to optimize complete iterative refinement rollouts from~\eqnref{iterative-cot} using Group Relative Policy Optimization (GRPO)~\cite{shao2024deepseekmath} with plan-level outcome supervision. In this section, we present a scalable reward design for open-world navigation and a tractable group-sampling procedure for assigning this reward to online plan-critique rollouts.

\paragraph{Reward design.}
It is challenging to hand-design rewards for open-world navigation as good plans must balance diverse constraints like task alignment, local safety, and long-range visual cues. We therefore learn a plan-quality reward from human preferences over candidate plans as shown in~\figref{trainingrecipe} section 2a. Given a context $x$ and pair of plans $(\zeta^+,\zeta^-)$, where $\zeta^+$ is preferred, we train a reward model $R_\phi(x,\zeta)$ with the Bradley-Terry~\cite{Bradley1952RankAO} objective:
\begin{equation}
\label{eq:rm-loss}
\mathcal{L}_{\mathrm{RM}}(\phi)
= - \mathbb{E}_{(x,\zeta^+,\zeta^-)}
\left[
\log \sigma\!\left(R_\phi(x,\zeta^+) - R_\phi(x,\zeta^-)\right)
\right].
\end{equation}
Following prior work on reward modeling~\cite{qiself, zhang2024grape}, we initialize $R_\phi$ from the finetuned VLM $\pi_\theta$ and replace the policy head with a linear reward head over the penultimate hidden state. Since $R_\phi$ scores only the context and plan, it is agnostic to the refinement history and critique text. We combine this learned reward with an expert-alignment reward $R_{\mathrm{exp}}(x,\zeta)$ that is inversely related to the Hausdorff distance between $\zeta$ and the expert plan $\hat{\zeta}$, yielding the final plan reward:
\begin{equation}
\label{eq:combined-reward}
R(x,\zeta)
=
R_\phi(x,\zeta)
+
\lambda R_{\mathrm{exp}}(x,\zeta),
\end{equation}
where $\lambda$ controls the strength of expert alignment. We define $R_{\mathrm{exp}}$ in Appendix~\ssecref{expertrewarddefinition}.

\paragraph{Practical group sampling and co-training.}
For each context $x$, we first sample a shared initial plan
$\zeta_0 \sim \pi^{\mathrm{plan}}_\theta(\cdot \mid x)$.
We then sample $G$ refinement rollouts conditioned on $\zeta_0$:
\[
\rho^i=(\zeta_0,z^i_0,\zeta^i_1,\ldots,\zeta^i_{\kappa_i},z^i_{\kappa_i}),
\qquad i=1,\ldots,G,
\]
where each rollout follows Eq.~\ref{eq:iterative-cot} after the initial plan. The rollout terminates when the critic accepts the current plan or when the refinement budget $K$ is reached; let $\kappa_i \le K$ denote this stopping step, and let $\zeta^i_{\kappa_i}$ denote the selected final plan. Using a shared $\zeta_0$ avoids confounding initial plan quality with the quality of subsequent critiques and refinements. We score each rollout using the outcome reward of its selected final plan and compute group-normalized advantages:
\begin{equation}
\label{eq:rollout-reward-advantage}
r^i = R(x,\zeta^i_{\kappa_i}),
\qquad
A^i =
\frac{
r^i-\mathrm{mean}_{j}(r^j)
}{
\mathrm{std}_{j}(r^j)+\epsilon
}.
\end{equation}
We optimize the shared VLM with Group Relative Policy Optimization (GRPO)~\cite{shao2024deepseekmath} using $A^i$ as the rollout-level advantage and a KL penalty to the frozen SFT reference policy $\pi_{\theta_{\mathrm{SFT}}}$. While this only provides indirect critique supervision~\cite{wang2024math}, we observe strong empirical gains and motivate this from an information-gain perspective in Appendix~\secref{appendix:reward-ig}, drawing parallels between how optimizing the critique for our proposed reward parallels discovering factors that lead to greater information gain.

\textbf{Approach Summary.} As mentioned at the beginning of this section and in~\figref{trainingrecipe}, our method consists of the iterative motion refinement formulation defined in~\eqnref{iterative-cot}, SFT pre-training to encourage expert-aligned plan-critique generation, and RL-post-training that uses a hybrid preference and expert alignment reward $R_\phi$ to provide outcome-level supervision. 

%% file: figures/trainingrecipe.tex
\begin{figure}[t]
    \vspace{-4mm}
    \centering
    \includegraphics[width=1.0\linewidth]{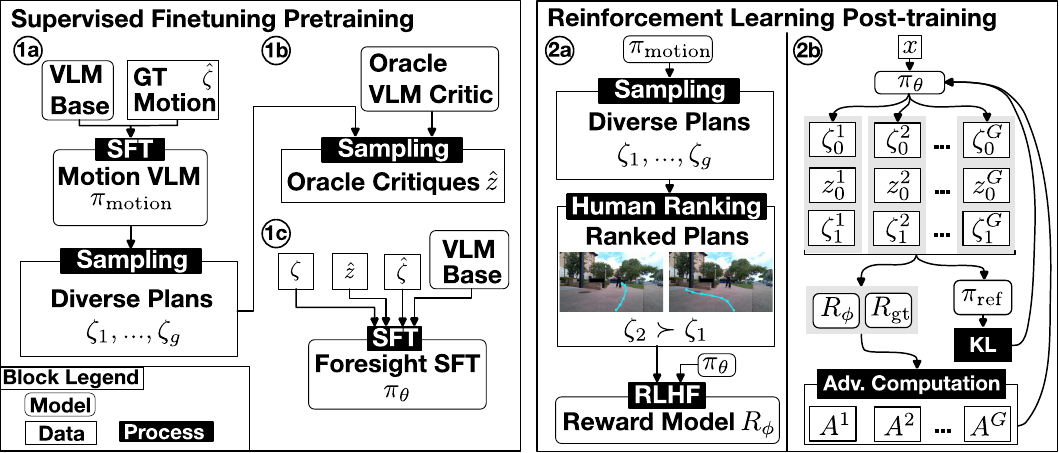}
    \caption{Overview of the \ourmodel{} training recipe. During supervised pre-training, we finetune a VLM for the iterative plan-critique (1c) roles using rollouts $(\zeta, \hat{z}, \hat{\zeta})$, the noisy plan, oracle critique, and ground truth plan respectively. In the second reinforcement learning stage, we learn a reward model for ranking motion plans $\zeta$ from a human-labeled preference dataset (2a) and optimize the VLM policy $\pi_\theta$ in the plan-critique loop using Group Relative Policy Optimization~\cite{shao2024deepseekmath} (2b).}
    \figlabel{trainingrecipe}
\end{figure}

%% file: experiments.tex
\section{Experiments and Results}
\seclabel{experiments}
\input{figures/testinglocations}

In this section, we describe the evaluation methodology for \ourmodel{} and answer the following questions to understand the importance of our contributions and overall performance on the task of mapless navigation with sparse language guidance.

\noindent
\hangindent=2.9em
\hangafter=1
\makebox[2.4em][l]{(\(\mathcal{Q}_1\))}
Does \ourmodel{} improve instruction-following navigation over state-of-the-art approaches, including feedforward or pre-emptive reasoning methods?

\noindent
\hangindent=2.9em
\hangafter=1
\makebox[2.4em][l]{(\(\mathcal{Q}_2\))}
Does reinforcement learning post-training improve \ourmodel{}'s ability to critique and refine plans around open-world navigation cues?

\noindent
\hangindent=2.9em
\hangafter=1
\makebox[2.4em][l]{(\(\mathcal{Q}_3\))}
Does \ourmodel{}'s proposed reward design improve refinement learning, or are verifiable learned or geometric rewards alone sufficient?
\vspace{0.0em}

We investigate these questions through offline and real-world robot experiments using state-of-the-art open and closed-source model baselines. For the offline benchmark, we collect an offline instruction following navigation dataset with expert demonstrations in a variety of urban environments.

\textbf{Experimental Setup.} We conduct all real-world tests using the Boston Dynamics Spot Robot using an Azure Kinect RGB-D camera to obtain RGB observations at 15Hz and odometry at 50Hz from the Boston Dynamics API. We use the same pure pursuit controller~\cite{huang2020path} and grounding model $\pi_\text{gnd}$ for baselines and provide additional architecture and training details in the Appendix~\ssecref{appendix:trainingdetailsgrounding}. For details on compute hardware and inference, please see Appendix~\ssecref{appendix:deploymentdetails}.

\textbf{Training and Evaluation Metholodogy.} We co-train all model baselines on SCAND~\cite{karnan2022socially} and 1.5 hours of teleoperated instruction following demonstrations (The \ourmodel{} Dataset). Additional dataset details, including qualitative samples, can be found in Appendix~\secref{appendix:foresightdatasetdetails}. 
To focus RL post-training on scenarios that require clue understanding, we use only the \ourmodel{} dataset and annotate 1 hour of demonstrations with ranked trajectory preferences. Additional methodological details are provided in Appendix~\ssecref{appendix:trainingdetailsreward}. We generate diverse language instructions, oracle textual critiques, and CoT reasoning traces using Gemini-3.1-Flash~\cite{pichai2025new}.

\input{figures/offlineexperiments_hausdorff}

For offline evaluation, we randomly withhold a subset of scenarios from the \ourmodel{} dataset to obtain 984 evaluation samples and use Hausdorff distance with the expert trajectory in bird's eye view (BEV) space. We use the same grounding policy $\pi_\text{gnd}$ to convert image space plans to BEV trajectories for consistency. In addition, we average 12 randomly generated rollouts for each test sample for more accurate performance estimatation. For more details on trajectory ranking, dataset examples, and prompts, please see Appendix~\secref{appendix:foresightdatasetdetails}.

For the real-world robot experiments, we evaluate in 6 environments categorized by the factor and navigation behavior being tested.~\figref{testinglocations} provides qualitative examples for the three categories: detour re-routing, structural clues, and sign understanding. We include 1 seen and unseen environment for each category and conduct 4 trials per baseline in each environment. We permit at most 3 interventions before deeming the test unsuccessful and only intervene when the robot becomes stuck, before catastrophic collisions, or after executing plans that make reaching the goal infeasible.

\input{tables/robotexperiments}

\textbf{Model Baselines.} We compare against several state-of-the-art language conditioned navigation baselines. We finetune LeLaN~\cite{hiroselelan}, a feedforward model pre-trained on internet navigation demonstrations. We reproduce  AlpaMayo~\cite{wang2025alpamayo}, a VLM that performs structured CoT reasoning about pre-defined factor classes and provides high-level directional guidance. To adapt Alpamayo for onboard reasoning in pedestrian friendly scenarios, we finetune Qwen3-VL-2B-Instruct~\cite{bai2025qwen3} to generate oracle CoT traces from Gemini-3.1-Flash and expert image space plans. Lastly, we compare against the closed-source Gemini-ER1.6~\cite{team2025gemini} VLM for offline evaluations, but do not include this for real-robot experiments due to internet connectivity and latency limitations. All \ourmodel{} models adapt the Qwen3-VL-2B-Instruct model for consistency. Additional baseline implementation details can be found in Appendix~\secref{appendix:trainingdetails}.

\subsection{Results and Analyses}

\textbf{Effectiveness of Iterative Plan-Critiques ($\mathcal{Q}_1$). } \figref{offlineexperiments} compares the performance of \ourmodel{} against other CoT baselines. Furthermore, it ablates the performance impact of iterative refinement and the relative importance of the motion and critic for downstream motion planning. Our approach trained only using SFT (\ourmodel{} SFT) outperforms the Gemini-ER1.6 and AlpaMayo baselines on average after 1 refinement iteration despite generating freeform reasoning traces, using fewer model parameters, and training on less data. Additionally, we observe that \ourmodel{} greatly outperforms a zero shot Qwen3VL model, corroborating our hypothesis that while VLMs contain some capacity for factor discovery and refinement, finetuning is necessary for understanding the most relevant factors and how they influence motion planning. We supplement this analysis with a real-world failure taxonomy for Alpamayo and \ourmodel{} in Appendix~\ssecref{appendix:ablationstudies-failuretaxonomy}. These results further support our claim that reasoning is both cue and plan dependent, highlighting how our method of jointly co-training the critic and planner reduces the percentage of critic failures by 23\% compared to pre-emptive CoT reasoning. For additional ablation studies on the relative importance of the motion planner versus critic, we refer readers to Appendix~\secref{appendix:ablationstudies}. We provide qualitative comparisons in Appendix~\ssecref{appendix:qualitativecomparison}.

\textbf{Effectiveness of RL Post-training ($\mathcal{Q}_2$).} Comparing RL post-training (RL/RL) to only SFT finetuning (SFT/SFT) in~\figref{offlineexperiments}, we find that RL post-training reduces the final planning error by 26\% on average. Our real-world experiments in~\tabref{robotexperiments} corroborate this finding, showing that RL improves the success rate by 20\% and decreases the number of interventions per mission by 36\% on average relative to SFT only (\ourmodel{}-rl). Thus, we conclude that our proposed RL post-training recipe is effective for improving real-world motion planning performance.

\textbf{Reward Design for RL Post-training ($\mathcal{Q}_3$).} 
\figref{offlineexperiments} compares different reward combinations used for RL post-training by measuring offline planning error with respect to expert demonstrations. 
While using only geometric rewards ($R_\text{exp}$) reduces planning error, it is substantially less effective than combining learned and geometric rewards (ours). 
In contrast, using the learned reward ($R_\phi$) alone is also ineffective. We hypothesize that learned and geometric rewards provide complementary signals that compensate for their individual weaknesses: geometric rewards do not capture non-geometric factors such as terrain or semantic affordances, leading to corner cutting and other undesirable behaviors, whereas learned rewards may capture acceptable trajectory modes without necessarily selecting the task-optimal plan. 
These findings suggest that combining learned and geometric rewards is important for stable RL convergence.

%% file: figures/testinglocations.tex
\begin{figure*}[htbp]
    \vspace{-1.0em}
    \centering
    \includegraphics[width=\linewidth]{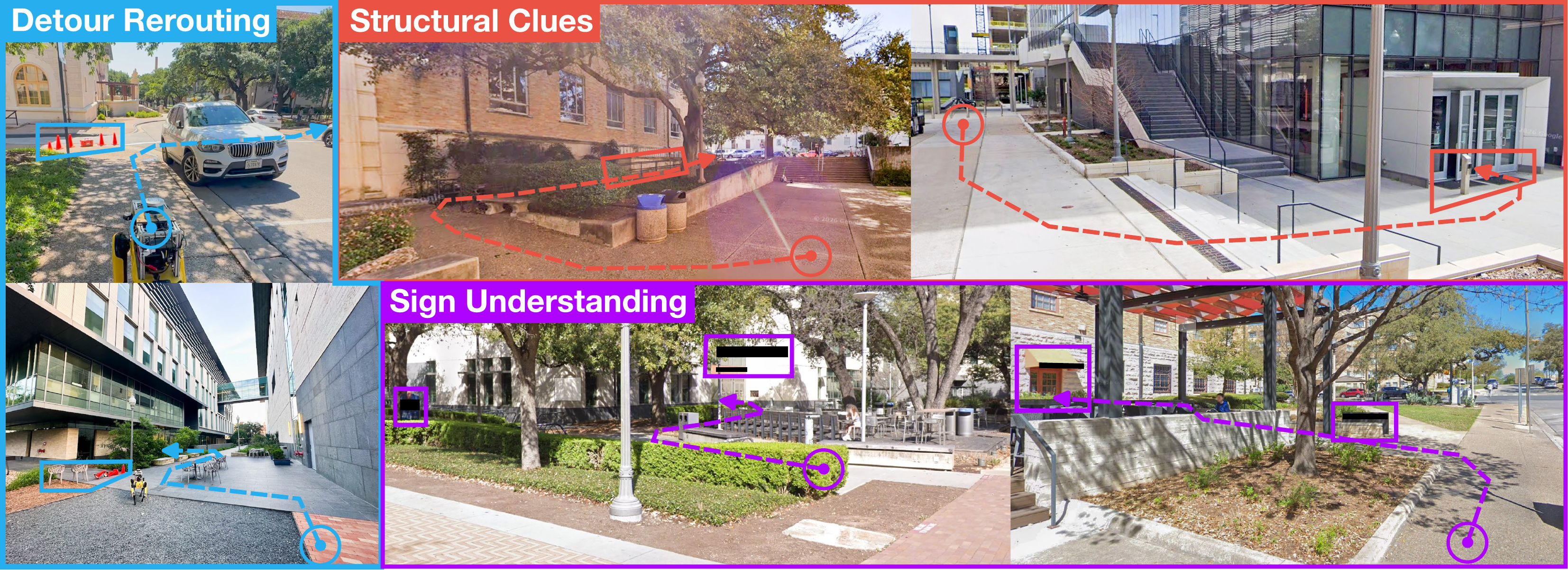}
    \caption{Real-world experiment scenarios. Bounding boxes  annotate the key visual clues for each scenario, and dashed arrows show the intended route from the start (circular dot). Annotations for visualization purposes only, not given to the algorithms. We redact building signs for anonymity.}
    \figlabel{testinglocations}
    \vspace{-1.0em}
\end{figure*}

%% file: figures/offlineexperiments_hausdorff.tex
\begin{figure*}
\centering
\includegraphics[width=1.0\linewidth]{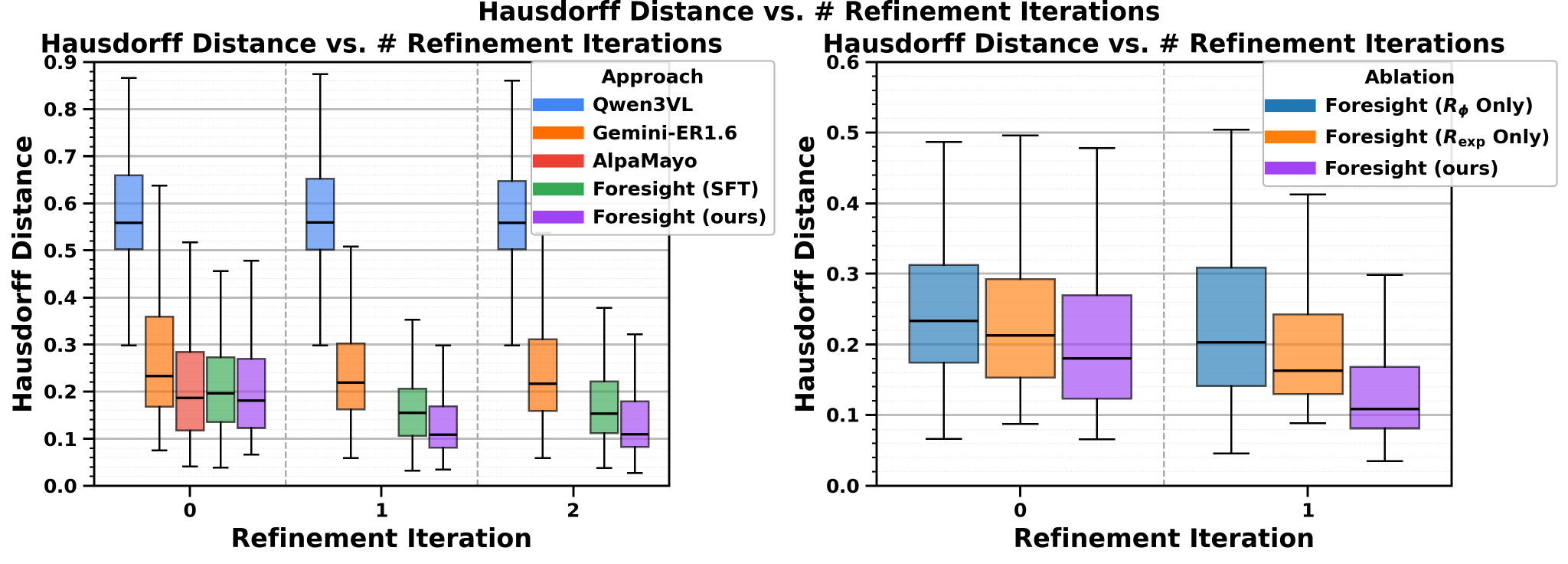}
\caption{Mean Hausdorff distance compared to expert demonstration with varying refinement iterations. The left plot compares \ourmodel{} against state-of-the-art baselines, where \ourmodel{} (SFT) is our approach but with only supervised pre-training. The right plot ablates different reward designs: $R_\phi$ - only using learned rewards, $R_\text{exp}$ - only using expert alignment reward. 
}
\figlabel{offlineexperiments}
\end{figure*}

%% file: tables/robotexperiments.tex
\begin{wraptable}[14]{r}{0.6\textwidth}
\centering
\scriptsize
\setlength{\tabcolsep}{2pt}
\renewcommand{\arraystretch}{0.9}
\resizebox{0.6\textwidth}{!}{%
\begin{tabular}{|l|cc|cc|cc|}
\hline
\multirow{2}{*}{\diagbox[height=3.3em,width=7.5em]{\textbf{Baseline}}{\textbf{Category}}}
& \multicolumn{2}{c|}{\textbf{Sign Under.}} 
& \multicolumn{2}{c|}{\textbf{Structural}} 
& \multicolumn{2}{c|}{\textbf{Detour}} \\ \cline{2-7}
& \makecell{\textbf{Succ.}\\\textbf{/ Trials}} 
& \makecell{\textbf{\#}\\\textbf{Int.}} 
& \makecell{\textbf{Succ.}\\\textbf{/ Trials}} 
& \makecell{\textbf{\#}\\\textbf{Int.}} 
& \makecell{\textbf{Succ.}\\\textbf{/ Trials}} 
& \makecell{\textbf{\#}\\\textbf{Int.}} \\ \hline
LeLaN~\cite{hiroselelan} & 1 / 8 & 3.00 & 2 / 8 & 2.00 & 2 / 8 & 2.00 \\ \hline
Alpamayo~\cite{wang2025alpamayo} & 4 / 8 & 2.00 & 3 / 8 & 1.67 & 4 / 8 & 2.00 \\ \hline
\ourmodel{}, no ref/rl & 2 / 8 & 2.00 & 4 / 8 & 2.00 & 3 / 8 & 2.33 \\ \hline
\ourmodel{}, no rl & 4 / 8 & 1.00 & 6 / 8 & 1.67 & 5 / 8 & 1.80 \\ \hline
\ourmodel{} (ours) & \textbf{7 / 8} & \textbf{0.71} & \textbf{7 / 8} & \textbf{1.14} & \textbf{6 / 8} & \textbf{1.00} \\ \hline
\end{tabular}%
}
\caption{Task success counts over total trials (Succ. / Trials) and \# of Interventions (\# Int.) across sign understanding (Sign Under.), structural clues, and detour rerouting real-world robot experiments. All baselines are allowed up to 3 interventions before failure. Here, no ref/rl indicates no reflections or RL finetuning.}
\tablabel{robotexperiments}
\vspace{-5pt}
\end{wraptable}

%% file: conclusion.tex
\section{Conclusion}
\seclabel{conclusion}

We presented \ourmodel{}, an iterative plan-critique framework for open-world mapless navigation from sparse language instructions. By adapting a VLM to critique and refine image-space plans, \ourmodel{} identifies novel plan-relevant visual factors and translates them into task-aligned motion updates. We train this loop with supervised pre-training and preference-based RL, enabling joint optimization of critique generation and critique-conditioned planning without dense critique-refinement annotations. Across offline and real-world experiments, \ourmodel{} improves task success by 37\% and reduces interventions per mission by 52\%, suggesting that iterative VLM self-critique is an effective mechanism for open-world robot motion refinement.

%% file: limitations.tex


\section{Limitations and Future Work}
\seclabel{limitations}

While \ourmodel{} makes significant strides towards scalable mapless navigation, several challenges remain. First, outcome-level supervision makes credit assignment difficult in multi-step refinement, since gains may arise from better critiques, critique-conditioned planning, or critique following. Process level supervision~\cite{shao2024deepseekmath} can provide denser supervision to mitigation spurious correlations and enhance instruction following. Second, limited memory and multi-view understanding can hurt long-horizon navigation when relevant cues are sparse or observed briefly. Alternate forms of memory like retrieval augmented generation~\cite{lewis2020retrieval} can provide more informative context and co-training on multi-view reasoning tasks~\cite{cheng20253d} can further improve multi-view cue grounding and reasoning. Finally, our experiments focus primarily on static environments, leaving dynamic settings that require faster policies and understand temporal relationships for future work.

%% file: appendix.tex
\clearpage
\appendix

\section{Appendix}

This appendix supplements the main paper with additional experimental analysis, reward derivations, dataset and prompt details, and model  implementation details. Specifically, we provide qualitative comparisons, quantitative ablations and a real-world failure taxonomy in~\secref{appendix:ablationstudies}; define the expert-alignment reward in~\secref{appendix:reward-design} and analyze its connection to critique-induced information gain in~\secref{appendix:reward-ig}; and describe the dataset, prompts, model training, reward learning, and deployment stack in~\secref{appendix:foresightdatasetdetails} and~\secref{appendix:trainingdetails}.

\section{Additional Results}
\seclabel{appendix:ablationstudies}

\subsection{Qualitative Comparisons}
\sseclabel{appendix:qualitativecomparison}

We provide additional qualitative comparisons between \ourmodel{} and the highest performing baseline, Alpamayo~\cite{wang2025alpamayo} in~\figref{qualitativecomparison}. We observe that Alpamayo is capable of inferring open-world visual clues like paths and doorway entrances, but often fails to focus on the most critical clues, like crosswalks, leading to suboptimal behavior compared to \ourmodel{}. Furthermore, we find that \ourmodel{} tends to refine the initial plan even if only small adjustments are needed. We hypothesize this occurs because reinforcement learning post-training optimizes our policy to make any improvements to the plan, regardless of their necessity. While this may improve robustness in scenarios with little margin for error, it may be potentially wasteful and motivates alternate reward designs that balance the inference cost of refinement with the current plan quality.

\input{figures/qualitativecomparison}

\input{figures/recipeablationexperiments}

\subsection{Training Recipe Ablations}
\sseclabel{appendix:ablationstudies-recipe}

In~\figref{recipeablationexperiments}, we conduct additional ablation sutdies to understand the importance of our training recipe decisions. Here, we adopt the naming convention A / B, where A represents the model used for motion planning and B represents the model used for the critic. For brevity, we use ZS to indicate that the model is tested Zero Shot, SFT to indicate that the model has undergone supervised finetuning (SFT), and RL to indicate that the model has undergone SFT and reinforcement learning (RL). We use Qwen3-VL-2B-Instruct as the base model for all ablations. 

We observe in~\figref{recipeablationexperiments} that all finetuned motion planners perform significantly better than zero shot. Furthermore, across iteration 0 (no refinements) to iteration 1, SFT/SFT outperforms SFT/ZS, demonstrating the importance of high quality critiques for motion refinement. Lastly, the planning error in ZS/SFT does not decrease with additional refinements, suggesting that simply finetuning the critic is not sufficient for motion refinement. We hypothesize this occurs because while the critiques may contain task-relevant factors for motion planning, the motion planner is unable to translate these factors to refine the motion plan. 

\subsection{Real-world Experiment Failure Taxonomy}
\sseclabel{appendix:ablationstudies-failuretaxonomy}

\input{figures/failuretaxonomy}

To better understand the causes real-world failure modes, we manually annotate the top-performing baseline, AlpaMayo~\cite{wang2025alpamayo} and \ourmodel{} to identify the categories of failures encountered. We order the diagram by if the cause of failure was due to poor CoT reasoning (critic) or motion planning (planner). Critic failures are either caused by ambiguous feedback or simply omitting the required factors needed for downstream planning. Planner failures are attributed to poor instruction following. In~\figref{failuretaxonomy}, we see that \ourmodel{} reduces the relative percentage of critic failures by 23\% from 81\% to 58\%. These results corroborate our claim that CoT reasoning is dependent on the visual cues and plans available to the policy. This also highlights the need for techniques that explicitly optimize for better motion planner instruction following and critics that understand not only what is in the scene, but what kind of feedback is helpful to guide motion planning.

\section{Reward Design}
\seclabel{appendix:reward-design}

\subsection{Expert-alignment Reward Definition}
\sseclabel{expertrewarddefinition}
Because our image-space plans are represented in normalized image coordinates, each waypoint lies in the unit square $[0,1]^2$. We measure the geometric distance between a predicted plan $\zeta$ and expert plan $\hat{\zeta}$ using the symmetric Hausdorff distance
\begin{equation}
d_{\mathrm{H}}(\zeta,\hat{\zeta})
=
\max\left\{
\sup_{p\in \zeta}\inf_{q\in \hat{\zeta}}\|p-q\|_2,
\;
\sup_{q\in \hat{\zeta}}\inf_{p\in \zeta}\|q-p\|_2
\right\}.
\end{equation}
Since both trajectories lie in $[0,1]^2$, the maximum possible pointwise distance is the unit-square diagonal $\sqrt{2}$. We convert this distance into a bounded reward by linearly mapping zero error to $1$ and a distance of $\sqrt{2}/2$ to $-1$, then clipping larger errors:
\begin{equation}
R_{\mathrm{exp}}(x,\zeta)
=
\mathrm{clip}\!\left(
1 - \frac{4}{\sqrt{2}} d_{\mathrm{H}}(\zeta,\hat{\zeta}),
-1,
1
\right).
\end{equation}
Thus, plans that closely match the expert receive high reward, while plans whose Hausdorff distance exceeds half the image diagonal receive the minimum reward.

\section{Reward Improvement and Information Gain}
\seclabel{appendix:reward-ig}

 
We derive the connection between the expected delta reward and the information gain induced by a critique. For a fixed context $x$, current trajectory $\zeta$, and critique $z$, define
\begin{equation}
p(\zeta) \;=\; \refp(\zeta \mid z, x),
\qquad
q(\zeta) \;=\; \planp(\zeta \mid x),
\end{equation}
so that $\IG(z) = \KL(p \,\|\, q)$ and $\dr(z) = \E_p[r] - \E_q[r]$. The Donsker--Varadhan~\citep{donsker} variational representation of KL divergence states
that for any function $f$ with $\E_q[e^{f}] < \infty$,
\begin{equation}
\KL(p \,\|\, q) \;\geq\; \E_p[f] - \log \E_q[e^{f}],
\label{eq:dv}
\end{equation}
with equality when $f = \log(p/q) + \mathrm{const}$. The inequality holds
for every admissible $f$, so we are free to choose any reward function $r$ and
obtain a valid lower bound.
 
Substituting $f = \lambda r$ with a free parameter $\lambda > 0$:
\begin{equation}
\IG(z) \;\geq\; \lambda \E_p[r] - \log \E_q[e^{\lambda r}].
\end{equation}
Decomposing $\E_p[r] = \dr(z) + \E_q[r]$ on the right-hand side:
\begin{equation}
\IG(z) \;\geq\; \lambda \, \dr(z) + \lambda \E_q[r] - \log \E_q[e^{\lambda r}].
\end{equation}
Equivalently,
\begin{equation}
\IG(z)
\geq
\lambda \dr(z) + C_q,
\qquad
C_q
=
\lambda \E_{\zeta \sim q}[r(\zeta)]
-
\log \E_{\zeta \sim q}[\exp(\lambda r(\zeta))].
\label{eq:appendix-delta-reward-ig}
\end{equation}

For a fixed policy snapshot, context $x$, and current trajectory $\zeta_t$, the baseline distribution $q$ is fixed when comparing different critiques $z_t$. Therefore, $C_q$ is constant with respect to the critique. Maximizing the expected
delta reward $\dr(z_t)$ therefore maximizes a Donsker--Varadhan lower bound on the information gain induced by the critique, under the current policy.

During training, the policy parameters change and therefore the baseline distribution $q$ also changes. We interpret Eq.~\ref{eq:appendix-delta-reward-ig} as a local justification for the reward objective: at each policy snapshot, critiques that yield larger expected reward improvement correspond to larger values of a lower bound on critique-induced information gain. 

\section{\ourmodel{} Dataset and Prompts}
\seclabel{appendix:foresightdatasetdetails}

In this section, we describe the environments of interest, oracle critique dataset generation procedure in ~\secref{appendix:oracle_critic_prompt}, and role-specific \ourmodel{} prompts in~\secref{appendix:prompts} used for training and inference.

\textbf{Environments of Interest.} The \ourmodel{} dataset consists of 88 missions in 25 unique urban/campus environments on a variety of goal-reaching navigation tasks that require sign understanding, inferring structural clues, and identifying detour routes.~\figref{trainingdataset} provides a gallery of scenarios with the natural language task, expert demonstration, and key visual clue highlighted for visualization purposes only. 

\input{figures/trainingdataset}

\subsection{Oracle Critique Dataset Generation Procedure.}
\seclabel{appendix:oracle_critic_prompt}

We prompt the oracle critic VLM (Gemini-3.1-Flash) by using a history of 4 image observations, annotating the last image with 5 motion plans sampled from our finetuned motion planner $\pi^\text{motion}_\theta$.~\figref{oraclecritiquecontext} shows example annotated images and the corresponding critic prompt below. 

\begin{figure*}[ht]
    \centering
    \includegraphics[width=\linewidth]{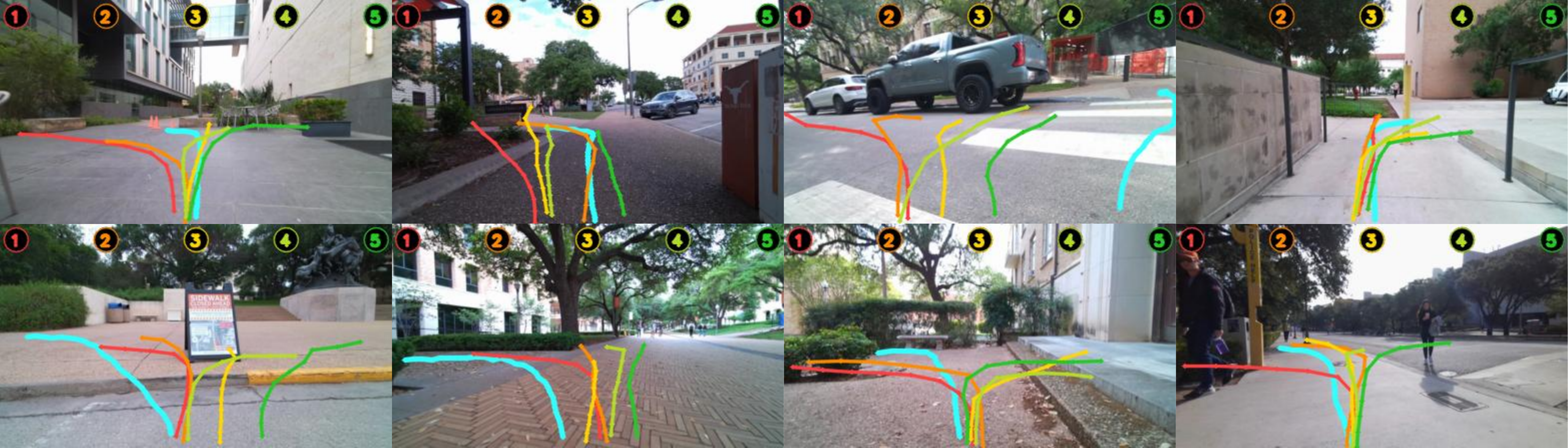}
    \caption{Qualitative examples of the the context images provided to the oracle critic VLM to use for generating critiques. Each motion plan is sampled from our motion planner and assigned an index. The critic VLM compared the plan with respect to the expert (cyan) and generates a short critique.}
    \figlabel{oraclecritiquecontext}
\end{figure*}

\input{prompts/criticgeneration}

\subsection{\ourmodel{} Prompts used for Training and Inference}
\seclabel{appendix:prompts}

We provide the exact prompts used for planning, critic, and refinement roles below in~\ssecref{appendix:motion_prompt}, \ssecref{appendix:motion_critic_prompt}, and \ssecref{appendix:motion_refinement_prompt} respectively.

\subsubsection{Motion Planning Prompt}
\sseclabel{appendix:motion_prompt}

\input{prompts/motiongeneration}

\subsubsection{Motion Critic Prompt}
\sseclabel{appendix:motion_critic_prompt}

\input{prompts/criticgeneration}
\subsubsection{Motion Refinement Prompt}
\sseclabel{appendix:motion_refinement_prompt}

\input{prompts/refinegeneration}

\section{Model Implementation Details}
\seclabel{appendix:trainingdetails}

In this section, we provide the grounding policy implementation details in~\ssecref{appendix:trainingdetailsgrounding}, supervised pre-training details in \ssecref{appendix:trainingdetailssft}, reward model training procedure in~\ssecref{appendix:trainingdetailsreward}, and supplementary deployment controller and inference latencies in~\ssecref{appendix:deploymentdetails}.

\subsection{Grounding Policy}
\sseclabel{appendix:trainingdetailsgrounding}

We closely follow prior work~\cite{zhang2026ventura} to implement our image-plan-conditioned grounding policy $\pi^\text{gnd}_\psi$. Let $o_t$ be the current image, $o_{\zeta_K}\in\mathbb{R}^{H\times W \times 1}$ be a boolean image annotated with the \ourmodel{} predicted plan, and $\zeta^\text{xyz} \in \mathbb{R}^{T \times 3}$ be a sequence of Cartesian XYZ waypoints. The grounding policy predicts Cartesian waypoints from the current observation and image-space plan as
\begin{equation}
\begin{aligned}
\pi^\text{gnd}_\psi(\zeta^\text{xyz} \mid o_t, o_{\zeta_K})
&=
g_\psi\!\left(
\operatorname{SSM}\!\left(
\left[
f^\text{depth}(o_t);
f^\text{plan}_\psi(o_{\zeta_K})
\right]
\right)
\right),
\end{aligned}
\eqnlabel{grounding_policy_arch}
\end{equation}
where $f^\text{depth}$ is a frozen DepthAnythingv2~\cite{yang2024depth} image encoder, $f^\text{plan}_\psi$ is a finetuned EfficientNet-B0~\cite{koonce2021efficientnet} encoder for the image-space plan, $[\cdot;\cdot]$ denotes channel-wise feature concatenation, $\operatorname{SSM}(\cdot)$ is a spatial softmax~\cite{finn2016deep}, and $g_\psi$ is a transformer encoder that maps the resulting latent features to Cartesian waypoints.

Concretely, we encode $o_t$ using the frozen DepthAnythingv2 encoder and $o_{\zeta_K}$ using the finetuned EfficientNet-B0 encoder. We stack the encoded features channel-wise and apply a spatial softmax to obtain latent features with dimension $K \times F$, where $K$ is the number of keypoints and $F$ is the feature dimension. Finally, we process the latent features with a transformer encoder to predict $\zeta^\text{xyz}$. We empirically determine that setting $K=384$, $F=384$, and the number of transformer encoder layers/heads to 8 yields the lowest average mean squared error (MSE) of 0.018m on the test set. This error indicates that the average error for each waypoint in the trajectory is less than 2 centimeters, which is acceptable for our motion planning domain.

We train $\pi^\text{gnd}_\psi$ on the SCAND and \ourmodel{} datasets to predict the expert Cartesian waypoints conditioned on the expert image-space plan $\hat{\zeta}^{\text{img}}$. We obtain $\hat{\zeta}^{\text{img}}$ by projecting the robot odometry into image space using known camera intrinsics and extrinsics.

\subsection{Supervised Pre-training}
\sseclabel{appendix:trainingdetailssft}

We initialize all \ourmodel{} models and the reproduced Alpamayo~\cite{wang2025alpamayo} model from Qwen3-VL-2B-Instruct. We summarize the supervise finetuning model, dataset, and optimization parameters for \ourmodel{} in~\tabref{sfthyperparameters}. While we allow up to 15 training epochs, we find that all models trained via SFT typically converge and begin overfitting within 8 training epochs.

\input{tables/sfthyperparameters}

\subsubsection{AlpaMayo Implementation Details}
\sseclabel{appendix:trainingdetailsalpamayo}

We train AlpaMayo on the same SFT dataset as \ourmodel{}, finetuning the base Qwen3vl model to predict a structured Chain-of-Causation (CoC) thinking trace first before predicting the image space motion plan. We use the following prompt for generating the oracle CoC traces using Gemini-3.1-Flash. We generate oracle traces for the same dataset split as was used for training the \ourmodel{} critic for fairness.

\input{prompts/oraclealpamayogeneration}

Then, we use the following prompt when finetuning the VLM to predict the CoC trace. We use the same motion prompt as \ourmodel{} for predicting the motion plan.

\input{prompts/alpamayogenerationprompt}

\subsubsection{LeLaN Implementation Details}

We initialize LeLaN using the open-source pre-trained checkpoint to ensure the base model contains internet-pretrained navigation knowledge. The original LeLaN model predicts a horizon of linear and angular velocities $[v, w]=\mathbb{R}^{T \times 2}$ and convert this to a sequence of Cartesian xy waypoints using Euler integration. We finetune the model end-to-end to predict linear and angular velocities on the same SCAND and \ourmodel{} dataset split as for \ourmodel{}. We also convert the model predictions to Cartesian xy waypoints using Euler integration and track these using the same low-level motion controller described in Appendix~\ssecref{appendix:deploymentdetails}.

\subsection{Preference Reward Learning}
\sseclabel{appendix:trainingdetailsreward}

In this section, we provide additional details for the trajectory annotation tool in~\ssecref{appendix:trajectoryrankingprocedure} and training procedure for the learned reward model $R_\phi$ in~\ssecref{appendix:rewardlearninghyperparameters}. 

\subsubsection{Trajectory Ranking Procedure.} 
\sseclabel{appendix:trajectoryrankingprocedure}

\figref{trajectoryrankingtool} shows the web tool used for ranking candidate motion plans. In the tool, the human annotator is shown the observation history, natural language task, and ground truth image plan. The annotator can use the Rollout button to sample sets of candidate plans from the finetuned VLM motion planner $\pi^\text{motion}_\theta$. Then, the annotator uses the dropdown to rank motion plans on a Likert scale from 1 (most preferred) to 5 (least preferred). For this work, we use a single human annotator and rank the trajectories relative to each other rather than using an absolute scale. We allow ranking ties, but do not use this for reward learning. In total, we annotate 1150 samples with 5 trajectories for each sample. 

\input{figures/trajectoryrankingtool}

\subsubsection{Reward Learning Details.} 
\sseclabel{appendix:rewardlearninghyperparameters}

Following prior work on finetuning VLMs as reward models~\cite{qiself}, we initialize a linear layer that predicts a scalar reward from the penultimate feature from the langauge model backbone. We initialize our reward model from $\pi_\theta$, our model that has already been supervise finetuned for planning, critiquing, and refinement. Empirically, we observe that initializing from the SFT-ed is necessary to prevent overfitting compared to using the base Qwen3-VL-2B-Instruct model. We prompt the reward model using the same context as the critic policy $\pi^\text{critic}_\theta$, which contains the predicted motion plan in the context history. 

\subsubsection{Reinforcement Learning Details.} 
\sseclabel{appendix:trainingdetailsrl}

We provide the model architecture, training hyperparameters, and dataset details in~\tabref{rlhyperparameters}. For this work, we limit the maximum number of refinement steps $K=1$ as we empirically observe in~\figlabel{recipeablationexperiments} that the model sees the most improvement on the first refinement round during supervised pre-training.

In regards to hyperparameter settings, we find that while the model is not particularly sensitive to the reward weighting parameters, the qualitative performance of the model is better when the reward weighting is biased more in factor of $R_\phi$ compared to $R_\text{exp}$. In addition, we observe that while accumulated reward continues to increase steadily without plateauing until roughly the end of epoch 7, the qualitative performance is best after 2-3 epochs of training. We hypothesize this occurs because we use the same motion planner for sampling the initial plan and for plan refinement. As a result, the motion planner begins to ignore the critiques later in training because the initial motion plan sampled does not require refinement. As mentioned in the limitations section, we acknowledge this is caused by a lack of credit assignment to incentivize following the critique and believe it can be addressed by computing auxiliary rewards to penalize ignoring the critique.

\input{tables/rlhyperparameters}

\subsection{Deployment Implementation Details}
\sseclabel{appendix:deploymentdetails}

This provides additional details on the compute hardware, inference latency benchmarking, and trajectory execution procedure.

We perform synchronous model inference and control solely using the onboard Nvidia AGX Orin (64GB) with a 12-core Arm Cortex A48AE CPU. On average across two environments, \ourmodel{} requires 3.05 seconds to do inference for each plan-critique loop, where 1.66 seconds are spent generating the plan and 1.39 seconds for the critique. Alpamayo takes 5.82 seconds to do inference, spending 1.77 seconds on planning and 4.05 seconds to generate the Chain-of-Causation reasoning trace. 

We standardize all model baselines to output a sequence of 10 Cartesian xyz waypoints. Because LeLaN originally predicts a horizon of linear and angular velocity commands, we integrate these predictions to obtain a trajectory and resample this trajectory to 10 waypoints. We track all waypoint plans using the same pure pursuit motion controller. Our navigation stack automatically follows each plan to the last waypoint, aligns the final heading with the heading angle between the penultimate and final waypoint, and re-queries the model for a new motion plan. This process is repeated until the robot reaches the goal or until the robot gets stuck and requires an intervention. For each intervention, we reset the robot to the closest nearby position and heading angle such that at least one visual clue is present for determining the future motion plan direction.

%% file: figures/qualitativecomparison.tex
\begin{figure*}[ht]
    \centering
    \includegraphics[width=\linewidth]{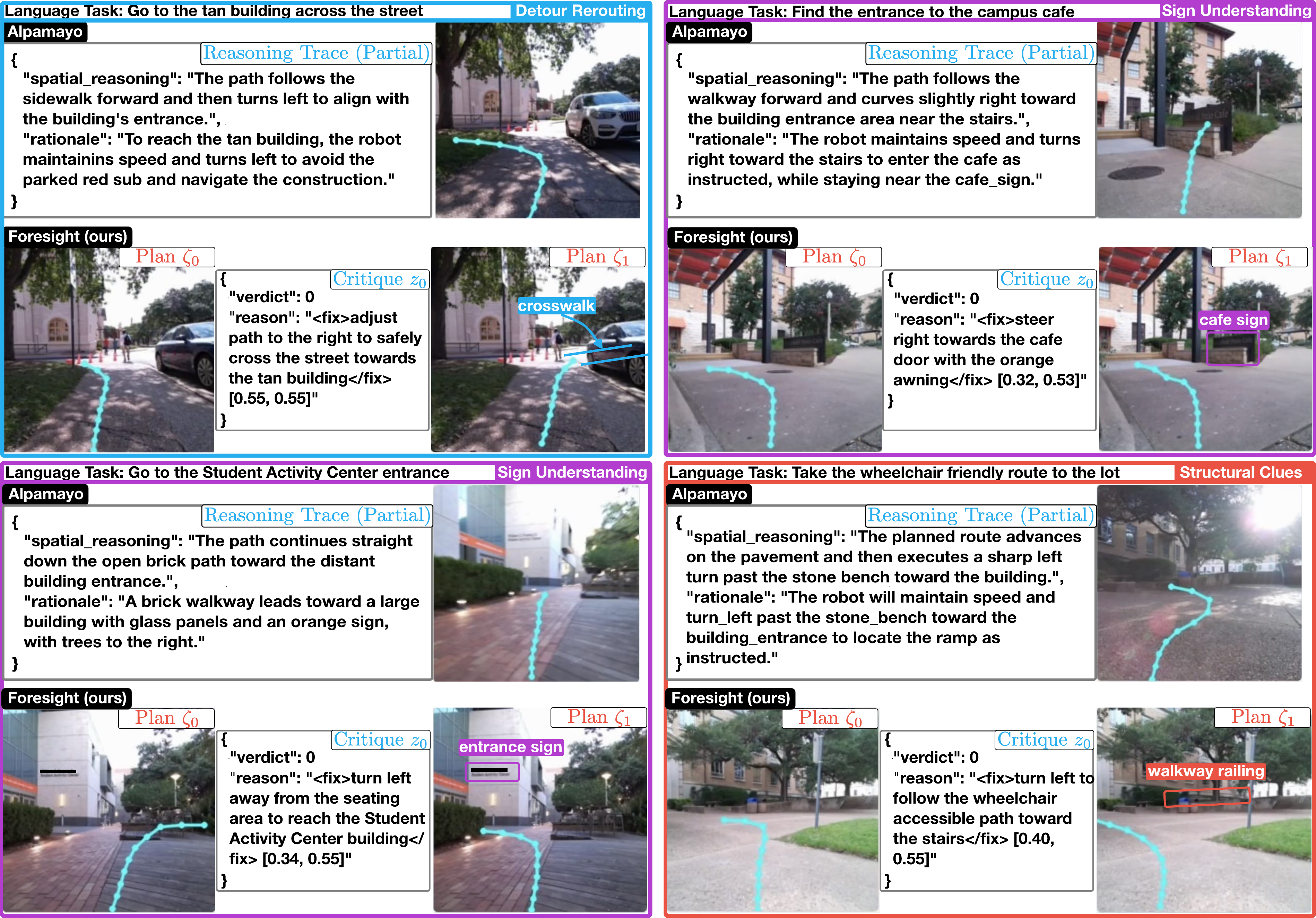}
    \caption{Qualitative comparison between \ourmodel{} and Alpamayo~\cite{wang2025alpamayo} across various experiments. We annotate the visual clue for visualization only.}
    \figlabel{qualitativecomparison}
    \vspace{-1.0em}
\end{figure*}

%% file: figures/recipeablationexperiments.tex
\begin{figure*}[htb]
\vspace{-10pt}
\centering
\includegraphics[width=0.7\linewidth]{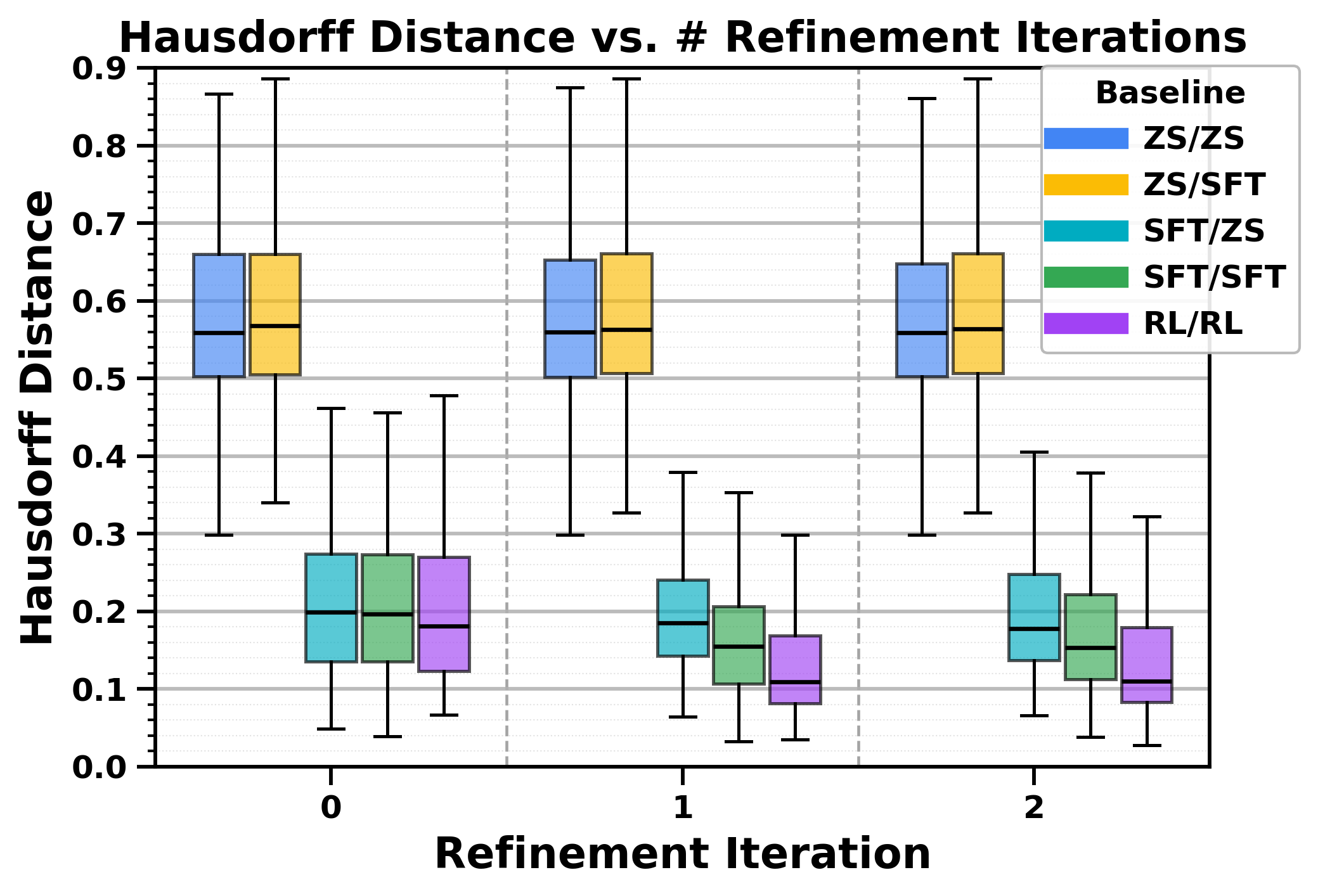}
\caption{Average Hausdorff distance error compared to expert demonstration with respect to the number of refinement iterations. For a full explanation of the naming convention used in the legend, we refer the reader to Appendix~\secref{appendix:ablationstudies}.}
\figlabel{recipeablationexperiments}
\end{figure*}

%% file: figures/failuretaxonomy.tex
\begin{figure*}[ht]
    \centering
    \includegraphics[width=\linewidth]{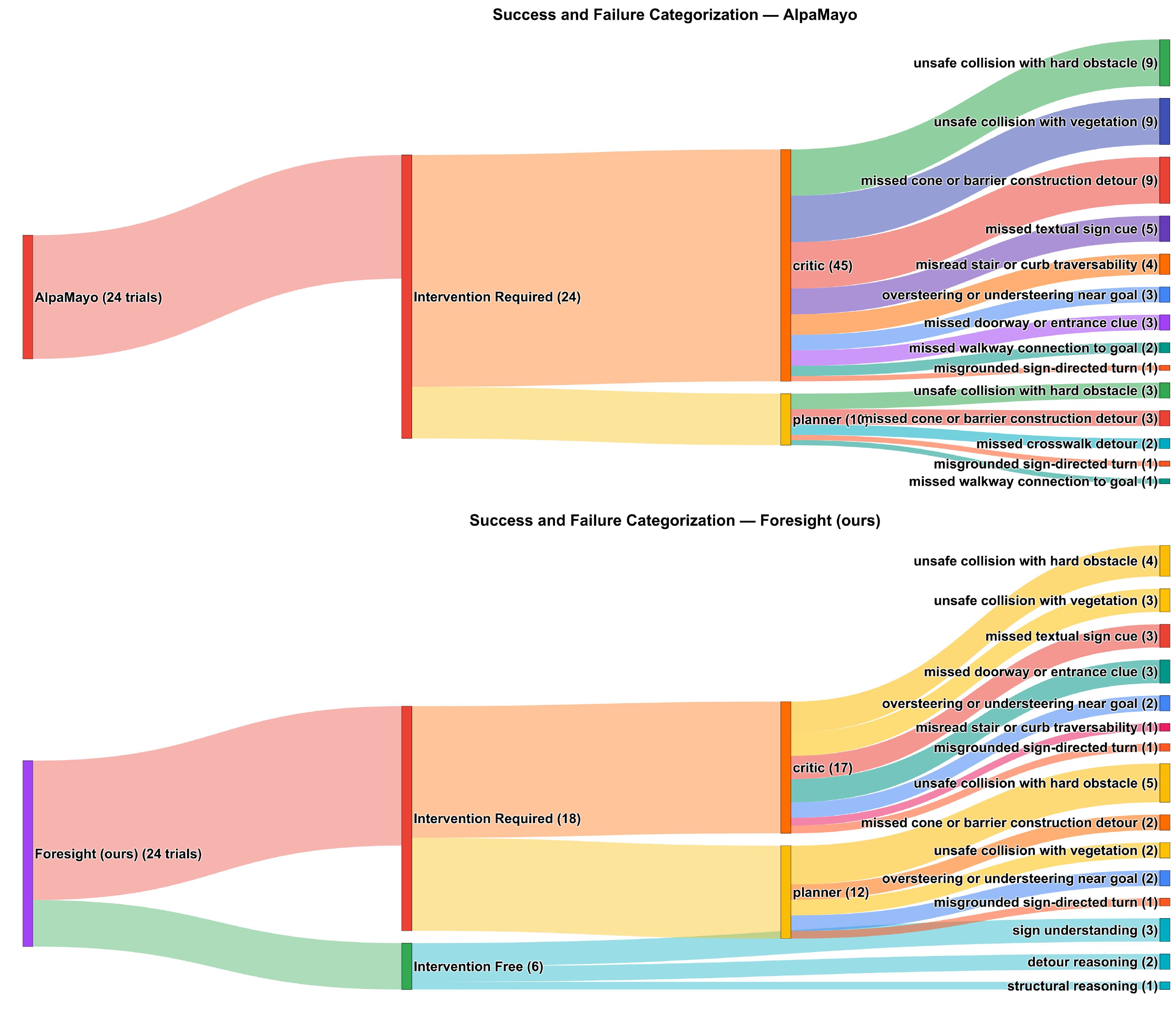}
    \caption{Success and Failure Taxonomy for Real World Experiments. We categorize the failures based on if they are caused by the critic or planner before describing the exact scene factor that caused each intervention.}
    \figlabel{failuretaxonomy}
\end{figure*}

%% file: figures/trainingdataset.tex
\begin{figure*}[htbp]
    \centering
    \includegraphics[width=1.0\linewidth]{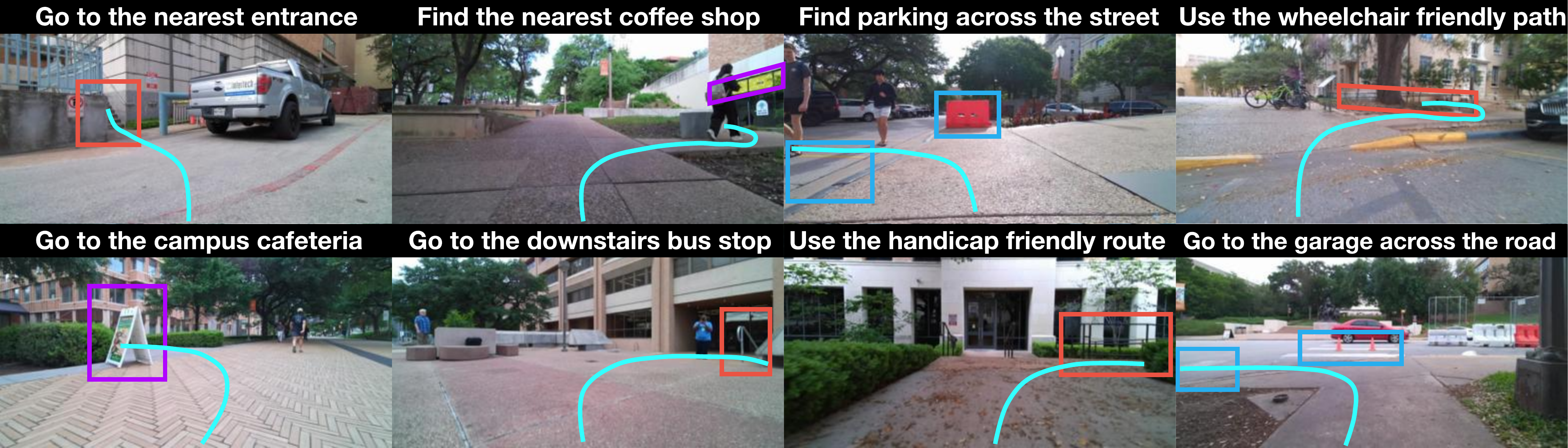}
    \caption{Examples from the \ourmodel{} dataset. For visualization purposes only, we highlight key visual clues for satisfying the language task using bounding boxes: Red for structural clues, purple for sign understanding, and blue for detours. The expert demonstration path is drawn in cyan.}
    \figlabel{trainingdataset}
    \vspace{-1.0em}
\end{figure*}

%% file: prompts/criticgeneration.tex
\begin{promptbox}
\noindent\textbf{Critic Prompt}

\begin{PromptVerbatim}
Attached is the egocentric robot image annotated with the planned path in yellow. The path coordinates are normalized xy path coordinates:
<|motion_start|><|motion_end|>

Think crucially about the language instruction and the path to analyze if the path is safe and appears to follow navigation cues that lead to the goal. If the path is acceptable, output "1" with the string "good" for the reason. If the path is unacceptable, output "0" and describe SPATIALLY AND SEMANTICALLY SPECIFIC VISIBLE CUES that are important to pay attention and how to improve them for key invalid points. The reason string should be like the following:
<fix> </fix> [x_{i}, y_{i}]
...
<fix> </fix> [x_{i+N}, y_{i+N}]

Critique points to consider: obstacle collisions, path plans that are likely to lead to the goal, traversable terrain for a pedestrian, or unrealistic robot navigation movements. Prescribe a direction to move in the correction.
Constraints: Ensure that no points are above the sky horizon line

OUTPUT RULES (MUST FOLLOW):
- Output ONLY valid JSON on ONE LINE.
- Output must start with { and end with }.
- Use exactly two keys: "verdict" and "reason".
- Do not critique more than two points
- Do NOT write "Verdict:" or "Reason:" or any extra text.

JSON template:
{"verdict":"0|1","reason":"<short image-grounded reason>"}

Generate the critique now.
\end{PromptVerbatim}
\end{promptbox}

%% file: prompts/motiongeneration.tex
\begin{promptbox}
\noindent\textbf{Motion Planning Prompt}

\begin{PromptVerbatim}
Attached are egocentric navigation images from a robot navigating to a goal. The images are in chronological order, where the last image is the current observation. Your task is to sample a unique motion trajectory from the distribution of trajectories that follows the language instruction:
(<|language_goal|>)
while satisfying the following constraints.

Constraints:
- The trajectory must be in normalized pixel coordinates. Each point is [x,y] with 0<=x<=1, 0<=y<=1.
- Use == 10 points. The first point MUST start near the bottom of the image.
- Points must on the walkable ground, but can be behind obstacles if the obstacle is passable (like a door or a wall).

Output ONLY JSON with exactly one key "trajectory". No extra text.
Format:
{"trajectory":[[x0, y0], ... [xn, yn]]}
\end{PromptVerbatim}
\end{promptbox}

%% file: prompts/refinegeneration.tex
\begin{promptbox}
\noindent\textbf{Motion Refinement Prompt}

\begin{PromptVerbatim}
Reflect on the previous planned path and the motion plan critique. If the verdict is 1, output the same path. If the verdict is 0, consider the issues mentioned in the critique to sample an improved motion plan that fixes valid issues while following the language instruction:
(<|language_goal|>)
while satisfying the earlier constraints:
- The trajectory must be in normalized pixel coordinates. Each point is [x,y] with 0<=x<=1, 0<=y<=1.
- Use == 10 points. The first point MUST start near the bottom of the image.
- Points must on the walkable ground, but can be behind obstacles if the obstacle is passable (like a door or a wall).

OUTPUT THE REFINED TRAJECTORY AS JSON IN THIS FORMAT:
{"trajectory":[[x0, y0], ... [xn, yn]]}
\end{PromptVerbatim}
\end{promptbox}

%% file: tables/sfthyperparameters.tex
\begin{table}[t]
\centering
\small
\caption{Supervised finetuning training parameters.}
\tablabel{sfthyperparameters}
\begin{tabular}{ll}
\toprule
\textbf{Category} & \textbf{Value} \\
\midrule
\multicolumn{2}{l}{\textit{Model}} \\
Base model & Qwen3-VL-2B-Instruct \\
Parameters & 1.8B total, 0.6M trainable \\
Trainable modules & LoRA on vision backbone, projector, and LLM \\
Planner/critic & Shared VLM backbone \\
\midrule
\multicolumn{2}{l}{\textit{LoRA}} \\
Rank / alpha / dropout & 64 / 64 / 0.0 \\
Target modules / bias & \texttt{all-linear} / none \\
\midrule
\multicolumn{2}{l}{\textit{Input/output}} \\
Image input & 4 frames at 224$\times$392 \\
Trajectory output & 10 normalized image-space waypoints in $[0,1]^2$ \\
Critic output & Binary verdict + free-form critique \\
Max sequence length & 4096 / 8192 tokens \\
Max output length & 192 motion tokens, 256 critic tokens \\
\midrule
\multicolumn{2}{l}{\textit{Dataset}} \\
Training / validation examples & 34,164 / 6,009 \\
SFT targets & Plans, critiques, and critique-conditioned refinements \\
\midrule
\multicolumn{2}{l}{\textit{Optimization}} \\
Optimizer / scheduler & AdamW / cosine decay \\
Learning rate / warmup / weight decay & $1\times10^{-4}$ / 0.0 / 0.01 \\
Global / per-device batch size & 512 / 256 \\
Max epochs / grad. clipping & 15 / 1.0 \\
Precision & bfloat16 \\
\midrule
\multicolumn{2}{l}{\textit{Compute}} \\
Hardware / time & 1$\times$ NVIDIA GH200 / $\sim$4 hours \\
Framework & Volcano Engine Reinforcement Learning (verl)~\cite{sheng2025hybridflow} \\
\bottomrule
\end{tabular}
\end{table}

%% file: prompts/oraclealpamayogeneration.tex
\begin{promptbox}
\noindent\textbf{Oracle Alpamayo CoC Thinking Prompt}

\begin{PromptVerbatim}
Attached are egocentric navigation images from a robot navigating to a goal. The images are in chronological order, where the last image is the current observation. The robot must follow the language instruction:
(<|language_goal|>)

The ground-truth motion plan has been drawn directly on the current observation as a cyan/teal polyline connecting the robot's planned waypoints from its current position toward the goal. Use this overlay as the authoritative reference for what the correct next action is: your reasoning must be consistent with and explain why this specific path was chosen given the scene. When generating the reasoning trace, do not mention the motion play overlay in your reasoning.

Generate a structured chain-of-causation reasoning trace that can be used to condition a subsequent motion plan, so it must commit to a concrete decision and clearly identify the causal factors driving it.

Field definitions:
- scene_caption: short caption describing traversable surfaces, doorways, dynamic obstacles, and lighting in the current observation.
- entities: grounded list of up to three distinct visual entities visible in the current image. Each entity has a "name" (snake_case noun) and a single "location" pixel point [x, y] in normalized coordinates with 0<=x<=1 and 0<=y<=1, marking the centroid of the entity.
- clues: chain-of-causation dictionary identifying which subset of entities causally drives the next decision. Keys MUST exactly match names from "entities". Values are short role descriptions (e.g., "target landmark for current subgoal", "dynamic obstacle, keep safe distance from"). List up to three clues.
- spatial_reasoning: one short sentence explaining the spatial relationship the agent must respect, grounded in the visible cyan motion-plan overlay (e.g., "the plan curves left around the pedestrian toward the open corridor entrance").
- meta_action: discrete decision pair with two keys: "longitudinal" in {stop, slow, maintain, accelerate} and "lateral" in {keep, turn_left, turn_right, sidestep_left, sidestep_right}. MUST be consistent with the direction of the drawn motion-plan polyline.
- rationale: one sentence linking the selected clues to the chosen meta_action causally and to the language instruction.

Constraints:
- Only reference entities visible in the current observation; do not hallucinate landmarks.
- Pixel locations must lie on the entity in the image and must not be above the sky horizon line. Each point is [x,y] with 0<=x<=1, 0<=y<=1.
- "clues" keys MUST be a subset of "entities" names.
- "rationale" MUST reference at least one clue, the chosen meta_action, the language instruction, and why the visible motion plan leads through the identified entities.
- Use SPATIALLY AND SEMANTICALLY SPECIFIC VISIBLE CUES; avoid vague descriptions such as "be cautious" or "watch out".
- Do not list weather, road type, or generic rule-based factors as causes unless they are directly visible and decision-driving.

OUTPUT RULES (MUST FOLLOW):
- Output ONLY valid JSON.
- Output must start with { and end with }.
- Use exactly six keys: "scene_caption", "entities", "clues", "spatial_reasoning", "meta_action", "rationale".
- Do NOT exceed three entries in "entities" or three entries in "clues".
- Do NOT write field labels, commentary, code fences, or any extra text outside the JSON.

JSON template:
{
  "scene_caption": "<short caption>",
  "entities": [
    {"name": "<snake_case>", "location": [x, y]}
  ],
  "clues": {
    "<entity_name>": "<causal role>"
  },
  "spatial_reasoning": "<one sentence>",
  "meta_action": {
    "longitudinal": "stop|slow|maintain|accelerate",
    "lateral": "keep|turn_left|turn_right|sidestep_left|sidestep_right"
  },
  "rationale": "<one sentence>"
}

Generate the reasoning trace now.
\end{PromptVerbatim}
\end{promptbox}

%% file: prompts/alpamayogenerationprompt.tex
\begin{promptbox}
\noindent\textbf{Training/Inference Alpamayo Thinking Prompt}

\begin{PromptVerbatim}
Attached are egocentric navigation images from a robot navigating to a goal. The images are in chronological order, where the last image is the current observation. The robot must follow the language instruction:
(<|language_goal|>)

Generate a structured chain-of-causation reasoning trace that can be used to condition a subsequent motion plan, so it must commit to a concrete decision and clearly identify the causal factors driving it.

Field definitions:
- scene_caption: short caption describing traversable surfaces, doorways, dynamic obstacles, and lighting in the current observation.
- entities: grounded list of up to three distinct visual entities visible in the current image. Each entity has a "name" (snake_case noun) and a single "location" pixel point [x, y] in normalized coordinates with 0<=x<=1 and 0<=y<=1, marking the centroid of the entity.
- clues: chain-of-causation dictionary identifying which subset of entities causally drives the next decision. Keys MUST exactly match names from "entities". Values are short role descriptions (e.g., "target landmark for current subgoal", "dynamic obstacle, keep safe distance from"). List up to three clues.
- spatial_reasoning: one short sentence explaining the spatial relationship the agent must respect (e.g., "hallway entrance is to the left, behind the pedestrian; wait for clearance before turning").
- meta_action: discrete decision pair with two keys: "longitudinal" in {stop, slow, maintain, accelerate} and "lateral" in {keep, turn_left, turn_right, sidestep_left, sidestep_right}.
- rationale: one sentence linking the selected clues to the chosen meta_action causally and to the language instruction.

Constraints:
- Only reference entities visible in the current observation; do not hallucinate landmarks.
- Pixel locations must lie on the entity in the image and must not be above the sky horizon line. Each point is [x,y] with 0<=x<=1, 0<=y<=1.
- "clues" keys MUST be a subset of "entities" names.
- "rationale" MUST reference at least one clue, the chosen meta_action, and the language instruction.
- Use SPATIALLY AND SEMANTICALLY SPECIFIC VISIBLE CUES; avoid vague descriptions such as "be cautious" or "watch out".
- Do not list weather, road type, or generic rule-based factors as causes unless they are directly visible and decision-driving.

OUTPUT RULES (MUST FOLLOW):
- Output ONLY valid JSON.
- Output must start with { and end with }.
- Use exactly six keys: "scene_caption", "entities", "clues", "spatial_reasoning", "meta_action", "rationale".
- Do NOT exceed three entries in "entities" or three entries in "clues".
- Do NOT write field labels, commentary, code fences, or any extra text outside the JSON.

JSON template:
{
  "scene_caption": "<short caption>",
  "entities": [
    {"name": "<snake_case>", "location": [x, y]}
  ],
  "clues": {
    "<entity_name>": "<causal role>"
  },
  "spatial_reasoning": "<one sentence>",
  "meta_action": {
    "longitudinal": "stop|slow|maintain|accelerate",
    "lateral": "keep|turn_left|turn_right|sidestep_left|sidestep_right"
  },
  "rationale": "<one sentence>"
}

Generate the reasoning trace now.
\end{PromptVerbatim}
\end{promptbox}

%% file: figures/trajectoryrankingtool.tex
\begin{figure*}[ht]
    \centering
    \includegraphics[width=\linewidth]{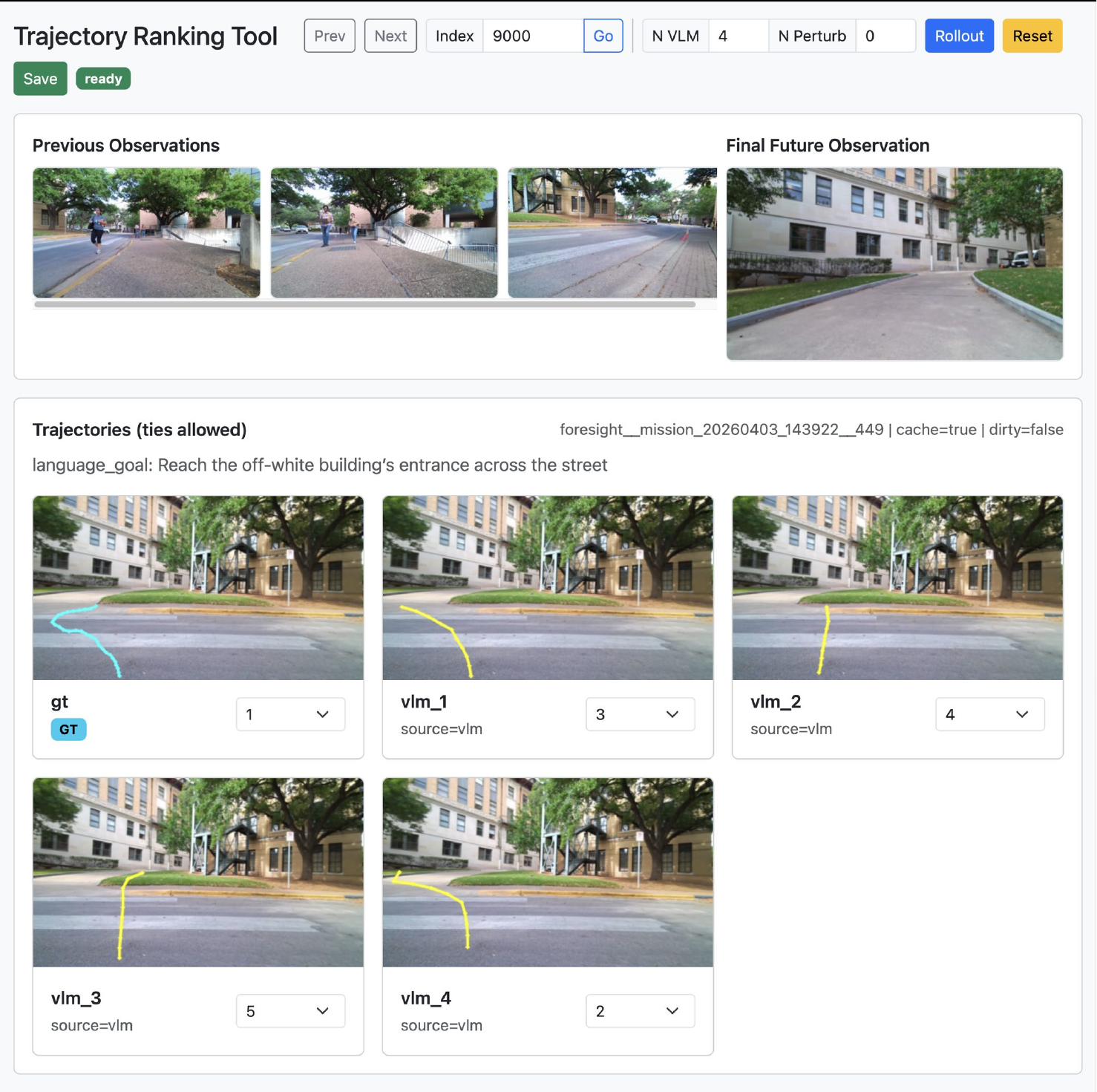}
    \caption{Web tool used for ranking motion plan candidates. The human annotator is shown the observation history, language goal, ground truth image plan (cyan), along with K alternate plans samples from our VLM motion planner.}
    \figlabel{trajectoryrankingtool}
    \vspace{-1.0em}
\end{figure*}

%% file: tables/rlhyperparameters.tex
\begin{table}[t]
\centering
\small
\caption{GRPO reinforcement learning training parameters.}
\tablabel{rlhyperparameters}
\begin{tabular}{ll}
\toprule
\textbf{Hyperparameter} & \textbf{Value} \\
\midrule
Initialization & SFT checkpoint~\ssecref{appendix:trainingdetailssft} \\
Trainable modules & LoRA on vision backbone, projector, and language backbone \\
Reference policy & Frozen SFT checkpoint~\ssecref{appendix:trainingdetailssft} \\
Reward model & Frozen preference reward model~\ssecref{appendix:trainingdetailsreward} \\
Algorithm & Group Relative Policy Optimization (GRPO) \\
Rollout unit & Initial Plan $\zeta_0$, Critique $z_0$, Refined Plan $\zeta_1$ \\
Group size $G$ & 8 rollouts per prompt \\
Max refinement steps $K$ & 1 \\
Reward weighting & 0.8 ($R_\phi)$, 0.2 ($R_\text{exp}$)  \\
Advantage normalization & Group-relative normalization \\
Loss aggregatation & Sequence mean, then token mean \\
KL coefficient $\beta$ & 0.01 \\
Clip range $\epsilon$ & 0.2 \\
Policy learning rate & $8\times10^{-5}$ \\
Optimizer / scheduler & AdamW / cosine decay \\
Global batch size & 64 prompts, 512 rollouts total \\
Mini-batch size & 16 \\
PPO/GRPO epochs & 10 \\
Sampling temperature & 1.0 \\
Top-$p$ & 0.95 \\
Max sequence length & 4096 / 8192 tokens \\
Max output length & 192 motion tokens, 256 critic tokens \\
Gradient clipping & 1.0 \\
Precision & bfloat16 \\
Hardware / time & 1$\times$ NVIDIA GH200 / 12 hours \\
Framework & Volcano Engine Reinforcement Learning (verl)~\cite{sheng2025hybridflow} \\
\bottomrule
\end{tabular}
\end{table}
